\documentclass[10pt]{article}

\usepackage{graphicx}
\usepackage{authordate1-4}
\begin{document}
\title{Computational models of attention}
\author{Laurent Itti and Ali Borji, University of Southern California}
\date{}
\maketitle

\section{Abstract}

This chapter reviews recent computational models of visual
attention. We begin with models for the bottom-up or stimulus-driven
guidance of attention to salient visual items, which we examine in
seven different broad categories. We then examine more complex models
which address the top-down or goal-oriented guidance of attention
towards items that are more relevant to the task at hand.

\section{Introduction}

A large body of psychophysical evidence on attention can be summarized
by postulating two forms of visual attention \cite{James1890}. The
first is driven by the visual input; this so-called exogenous,
bottom-up, stimulus-driven, or saliency-based form of attention is
rapid, operates in parallel throughout the entire visual field, and
helps mediate pop-out, the phenomenon by which some visual items tend
to stand out from their surroundings and to instinctively grab our
attention \cite{Treisman_Gelade80,Koch_Ullman85,Itti_Koch01nrn}. The
second, endogenous, top-down, or task-driven form of attention,
depends on the exact task at hand and on subjective visual experience,
takes longer to deploy, and is volitionally controlled. Normal vision
employs both processes simultaneously, to control both overt and
covert shifts of visual attention \cite{Itti_Koch01nrn}. Covert focal
attention has been described as a rapidly shiftable ``spotlight''
\cite{Crick84}, which serves the double function of selecting
particular locations and objects of interest, and of enhancing visual
processing at those locations and for specific attributes of those
objects. Thus, attention acts as a shiftable information processing
bottleneck, allowing only objects within a circumscribed visual region
to reach higher levels of processing and visual awareness
\cite{Crick_Koch98}.

Most computational models of attention to date have focused on
bottom-up guidance of attention towards salient visual items. Many,
but not all, of these models have embraced the concept of a
topographic saliency map \cite{Koch_Ullman85} that highlights scene
locations according to their relative conspicuity or salience. This
has allowed for validation of saliency models against eye movement
recordings, by measuring the extent to which human or monkey observers
fixate locations that have higher predicted salience than expected by
chance \cite{Parkhurst_etal02,Tatler_etal05}.  However, recent
behavioral studies have shown that the majority of eye fixations
during execution of many tasks are directed to task-relevant locations
that may or may not also be salient, and fixations are coupled in a
tight temporal relationship with other task-related behaviors such as
reaching and grasping \cite{hayhoe2003visual}.  Several of these
studies have used naturalistic interactive or immersive environments
to give high-level accounts of gaze behavior in terms of objects,
agents, ``gist of the scene'' \cite{Potter_Levy69,Torralba03}, and
short-term memory, to describe, for example, how task-relevant
information guides eye movements while subjects make a sandwich
\cite{Land_Hayhoe01,hayhoe2003visual}, or how distractions such as
setting the radio or answering a phone affect eye movements while
driving \cite{Sodhi_etal02}. While these more complex attentional
behaviors have been more difficult to capture in computational models,
we review below several recent efforts that have successfully modeled
attention guidance during complex tasks, such as driving a car or
running a hot-dog stand that serves many hungry customers.

While we focus on purely computational models (which autonomously
process visual data without the requirement of a human operator, or of
manual parsing of the data into conceptual entities), we also point
the reader to several relevant previous reviews on attention theories
and models more generally
\cite{Itti_Koch01nrn,Paletta_etal05,Frintrop_etal10,Gottlieb_Balan10,Toet11,Tsotsos_Rothenstein11}.

\section{Computational models of bottom-up attention}

\begin{figure}[p]
\includegraphics[width=\linewidth]{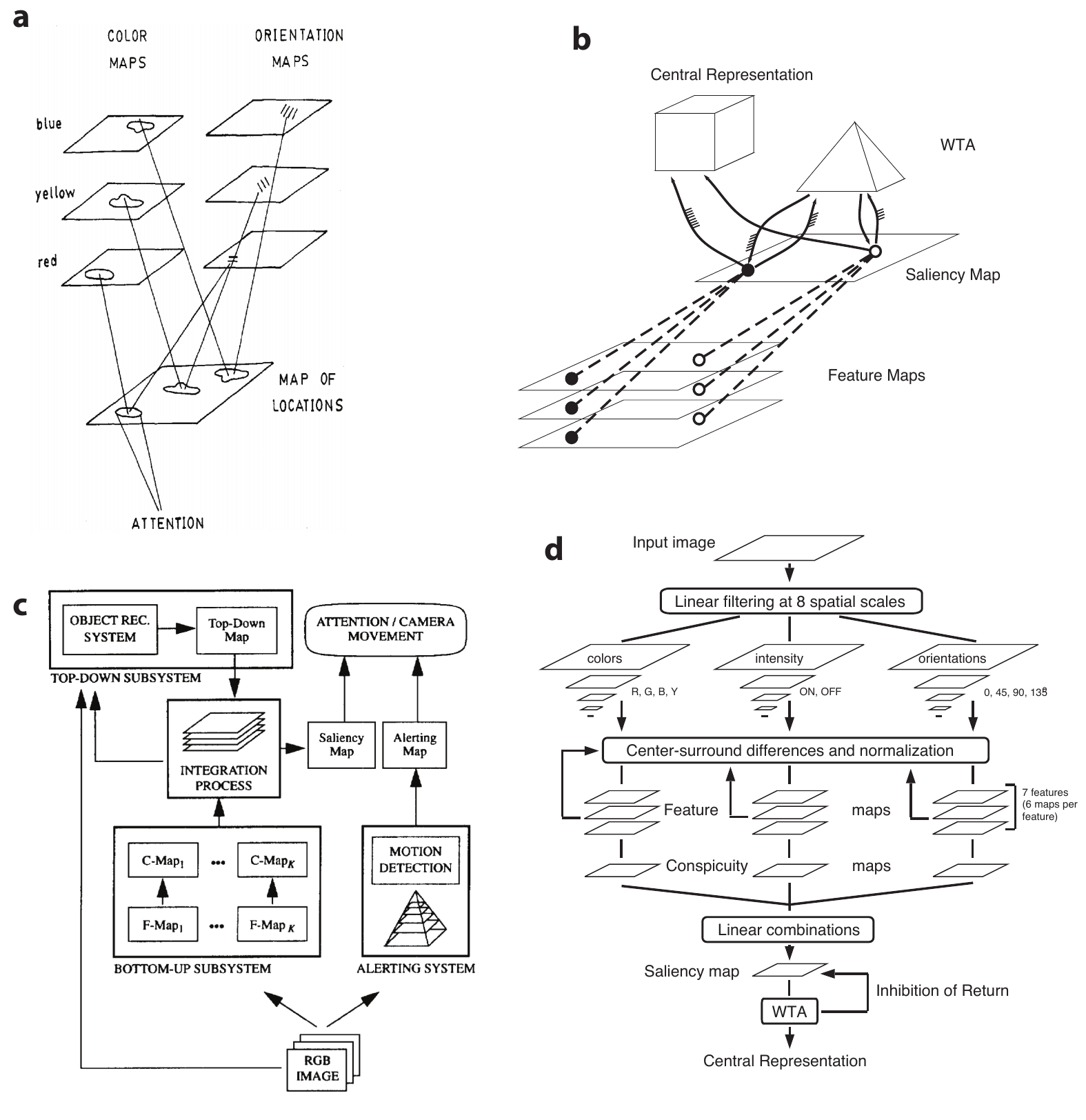}
\caption{Early bottom-up attention theories and models. {\bf (a)}
  Feature integration theory of Treisman \& Gelade (1980) posits
  several feature maps, and a focus of attention that scans a map of
  locations and collects and binds features at the currently attended
  location. (from Treisman \& Souther (1985)). {\bf (b)} Koch \&
  Ullman (1985) introduced the concept of a saliency map receiving
  bottom-up inputs from all feature maps, where a winner-take-all
  (WTA) network selects the most salient location for further
  processing. {\bf (c)} Milanese {\em et al.} (1994) provided one of
  the earliest computational models. They included many elements of
  the Koch \& Ullman framework, and added new components, such as an
  alerting subsystem (motion-based saliency map) and a top-down
  subsystem (which could modulate the saliency map based on memories
  of previously recognized objects). {\bf (d)} Itti {\em et al.}
  (1998) proposed a complete computational implementation of a purely
  bottom-up and task-independent model based on Koch \& Ullman's
  theory, including multiscale feature maps, saliency map,
  winner-take-all, and inhibition of return.} \label{FIGearlybu}
\end{figure}

Development of computational models of attention started with the
Feature Integration Theory of Treisman \& Gelade
\shortcite{Treisman_Gelade80}, which proposed that only simple visual
features are computed in a massively parallel manner over the entire
visual field.  Attention is then necessary to bind those early
features into a united object representation, and the selected bound
representation is the only part of the visual world that passes though
the attentional bottleneck (Figure~\ref{FIGearlybu}.a). Koch and
Ullman \shortcite{Koch_Ullman85} extended the theory by proposing the
idea of a single topographic {\em saliency map}, receiving inputs from
the feature maps, as a computationally efficient representation upon
which to operate the selection of where to attend next: A simple
maximum-detector or {\em winner-take-all} neural network was proposed
to simply pick the next most salient location as the next attended
one, while an active {\em inhibition-of-return} mechanism would later
inhibit that location and thereby allow attention to shift to the next
most salient location (Figure~\ref{FIGearlybu}.b). From these ideas, a
number of fully computational models started to be developed (e.g.,
Figure~\ref{FIGearlybu}.c,d).

\begin{figure}[p]
\includegraphics[width=\linewidth]{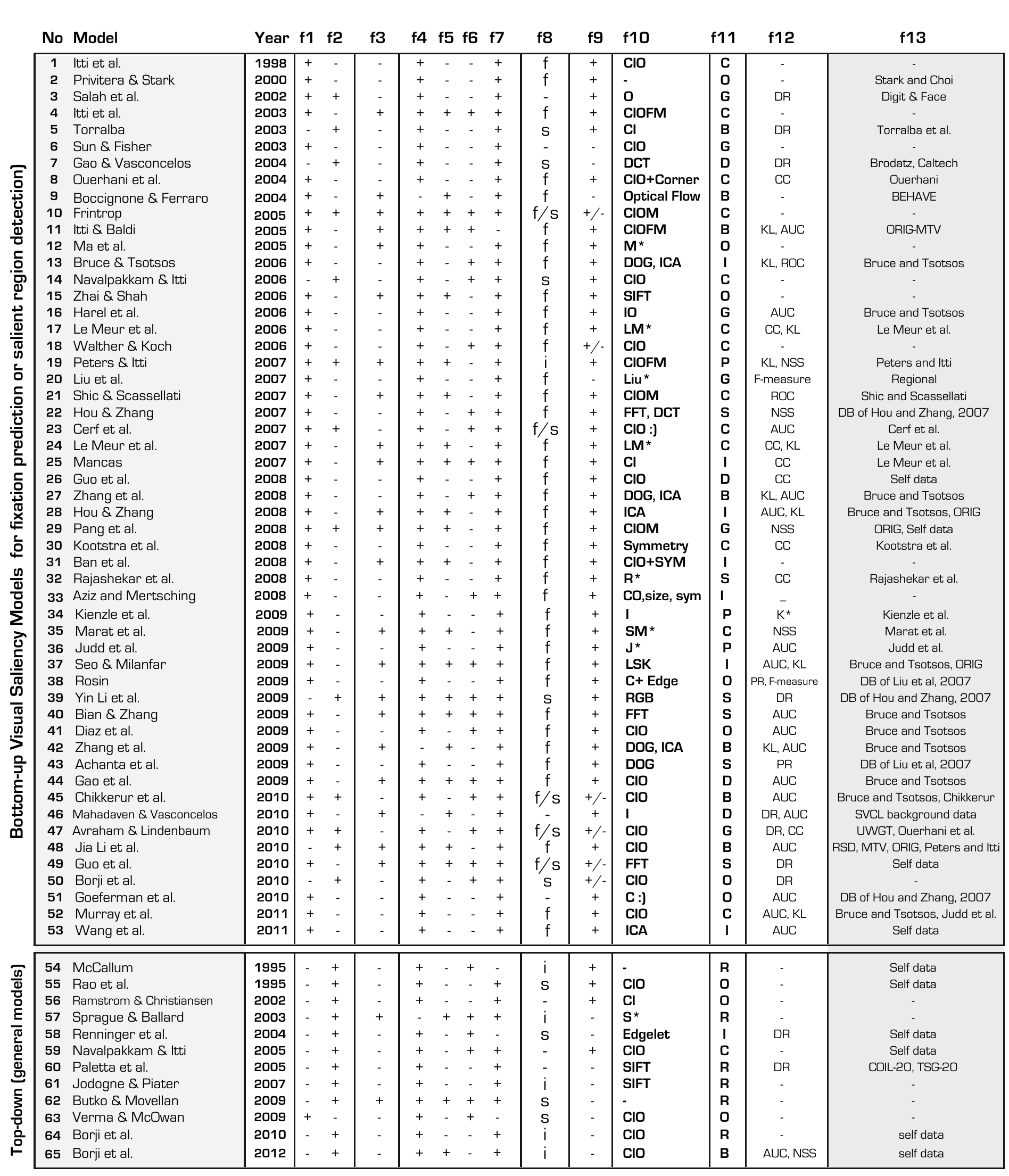}
\caption{\footnotesize Survey of bottom-up and top-down computational
  models, classified according to 13 factors. Factors in order are:
  Bottom-up ($f_{1}$), Top-down ($f_{2}$), Spatial (-)/Spatio-temporal
  (+) ($f_{3}$), Static ($f_{4}$), Dynamic ($f_{5}$), Synthetic
  ($f_{6}$) and Natural ($f_{7}$) stimuli, Task-type ($f_{8}$),
  Space-based(+)/Object-based(-) ($f_{9}$), Features ($f_{10}$), Model
  type ($f_{11}$), Measures ($f_{12}$), and Used dataset
  ($f_{13}$). In Task type ($f_{8}$) column: free-viewing ($f$);
  target search ($s$); interactive ($i$). In Features ($f_{10}$)
  column: CIO: color, intensity and orientation saliency; CIOFM: CIO
  plus flicker and motion saliency; M* = motion saliency, static
  saliency, camera motion, object (face) and aural saliency
  (Speech-music); LM* = contrast sensitivity, perceptual
  decomposition, visual masking and center-surround interactions; Liu*
  = center-surround histogram, multi-scale contrast and color
  spatial-distribution; R* = luminance, contrast, luminance-bandpass,
  contrast-bandpass; SM* = orientation and motion; J* = CIO,
  horizontal line, face, people detector, gist, etc; S* = color
  matching, depth and lines; :) = face. In Model type ($f_{11}$)
  column, R means that a model is based RL. In Measures ($f_{12}$)
  column: K* = used Wilcoxon-Mann-Whitney test (The probability that a
  random chosen target patch receives higher saliency than a randomly
  chosen negative one); DR means that models have used a measure of
  detection/classification rate to determine how successful was a
  model. PR stands for Precision-Recall. In dataset ($f_{13}$) column:
  Self data means that authors gathered their own data. For a detailed
  definition of these factors please refer to Borji \& Itti (2012
  PAMI).}
\label{FIGmodels}
\end{figure}

Many research groups have more recently developed new models of
bottom-up attention. Fifty three bottom-up models are classified along
13 different factors in Figure~\ref{FIGmodels}.  Most of these models
fall into one of the seven general categories described below, with
some models spanning several categories (also see
\cite{Tsotsos_Rothenstein11} for another taxonomy). Additional details
and benchmarking of these models was recently proposed by Borji {\em
  et al.}  \shortcite{Borji_Itti12pami,Borji_etal12tip}.

\textbf{Cognitive models}. Development of saliency-based models
escalated after Itti {\em et al.}'s~\shortcite{Itti_etal98pami}
implementation of Koch and Ullman's~\shortcite{Koch_Ullman85}
computational architecture. Cognitive models were the first to
approach algorithms for saliency computation that could apply to any
digital image. In these models, the input image is decomposed into
feature maps selective for elementary visual attributes (e.g.,
luminance or color contrast, motion energy), at multiple spatial
scales.  The feature maps are combined across features and scales to
form a master saliency map.  An important element of this theory is
the idea of center-surround operators, which define saliency as
distinctiveness of an image region compared to its surroundings.
Almost all saliency models are directly or indirectly inspired by
cognitive concepts of visual attention
(e.g.,~\cite{le2006coherent,MaratIJCV}).

\textbf{Information-theoretic models}. Stepping back from
biologically-plausible implementations, models in this category are
based on the premise that localized saliency computations serve to
guide attention to the most informative image regions first.  These
models thus assign higher saliency to scene regions with rare (low
probability) features.  Information of visual feature $F$ is $I(F)= -
log \mbox{ $p(F)$}$, inversely proportional to the likelihood of
observing $F$.  By fitting a distribution $P(F)$ to features, rare
features can be immediately found by computing $P(F)^{-1}$ at every
location in an image.  While, in theory, using any feature space is
feasible, often these models (inspired by efficient coding in visual
cortex) utilize a sparse set of basis functions learned from natural
scenes. Example models in this category are AIM~\cite{BruceNIPS},
Rarity~\cite{mancas2007computational}, LG (Local + Global image patch
rarity)~\cite{Borji_Itti12cvpr}, and incremental coding length
models~\cite{HouZhangNIPS2008}.

\textbf{Graphical models}. Graphical models are generalized Bayesian
models, which have been employed for modeling complex attention
mechanisms over space and time. Torralba~\shortcite{Torralba03}
proposed a Bayesian approach for modeling contextual effects on visual
search which was later adopted in the SUN
model~\cite{ZhangTong2Etal008} for fixation prediction in free
viewing. Itti \& Baldi~\shortcite{Itti_Baldi05cvpr} defined surprising
stimuli as those which significantly change beliefs of an
observer. Harel {\em et al.} \shortcite{harel2007graph} propagated
similarity of features in a fully connected graph to build a saliency
map.  Avraham \& Lindenbaum~\shortcite{Avraham_Lindenbaum10}, Jia
Li~{\em et al.},~\shortcite{li2010optimol}, and Tavakoli~{\em et
  al.}~\shortcite{rezazadegan2011fast}, have also exploited Bayesian
concepts for saliency modeling.

\textbf{Decision theoretic models}. This interpretation proposes that
attention is driven optimally with respect to the task. Gao \&
Vasconcelos~\shortcite{gao2004discriminant} argued that, for
recognition of objects, salient features are those that best
distinguish a class of objects of interest from all other
classes. Given some set of features $X = \{X_{1}, \cdots,X_{d}\}$, at
locations $l$, where each location is assigned a class label $Y$
($Y_{l} = 0$ for background, $Y_{l} = 1$ for objects of interest),
saliency is then a measure of mutual information (usually the
Kullback-Leibler divergence), computed as $I(X,Y) = \sum_{i=1}^{d}
I(X_{i}, Y)$. Besides having good accuracy in predicting eye
fixations, these models have been very successful in computer vision
applications (e.g., anomaly detection and object tracking).

\textbf{Spectral analysis models}. Instead of processing an image in
the spatial domain, these models compute saliency in the frequency
domain. Hou \& Zhang~\shortcite{HouZhangCVPR2007} derive saliency for
an image by computing its Fourier transform, preserving the phase
information while discarding most of the amplitude spectrum (to focus
on image discontinuities), and taking the inverse Fourier transform to
obtain the final saliency map.  Bian \& Zhang
\shortcite{BianZhang2009} and Guo \& Zhang \shortcite{GuoIEEEIP}
further proposed spatio-temporal models in the spectral domain.

\textbf{Pattern classification models}. Models in this category use
machine learning techniques to learn stimulus-to-saliency mappings,
from image features to eye fixations. They estimate saliency $s$ by
computing $p(s|f)$, where $f$ is a feature vector which could be the
contrast of a location compared to its surrounding
neighborhood. Kienzle~{\em et
  al.}~\shortcite{kienzle2007nonparametric}, Peters \&
Itti~\shortcite{Peters_Itti07cvpr}, and Judd~{\em et
  al.}~\shortcite{Judd_etal09} used image patches, scene gist, and a
vector of several features at each pixel, respectively, and used
pattern classifiers to then learn saliency from the features.
Tavakoli~{\em et al.}~\shortcite{rezazadegan2011fast} used sparse
sampling and kernel density estimation to estimate the above
probability in a Bayesian framework.  Note that some of these models
may not be purely bottom-up since they use features that guide
top-down attention, for example faces or text
\cite{Judd_etal09,Cerf_etal08}.

\textbf{Other models}. Other models exist that do not easily fit into
our categorization. For example, Seo \& Milanfar
\shortcite{SeoMilanfar2009JV} proposed self-resemblance of local image
structure for saliency detection. The idea of decorrelation of neural
response was used for a normalization scheme in the Adaptive Whitening
Saliency (AWS) model~\cite{garcia2009decorrelation}. Kootstra~{\em et
  al.}~\shortcite{kootstra2008paying} developed symmetry operators for
measuring saliency and Goferman~{\em et al.}
\shortcite{goferman2010context} proposed a context-aware saliency
detection model with successful applications in re-targeting and
summarization.

In summary, modeling bottom-up visual attention is an active research
field in computational neuroscience and machine vision. New theories
and models are constantly proposed which keep advancing the state of
the art.

\section{Top-down attention models} \label{SECtd}

Models that address top-down, task-dependent influences on attention
are more complex, as some representations of goal and of task become
necessary. In addition, top-down models typically involve some degree
of cognitive reasoning, not only attending to but also recognizing
objects and their context, to incrementally update the model's
understanding of the scene and to plan the next most task-relevant
shift of attention
\cite{Navalpakkam_Itti05vr,Yu_etal08,Beuter_etal09,Yu_etal12}.  For
example, one may consider the following information flow, aimed at
rapidly extracting a task-dependent compact representation of the
scene, that can be used for further reasoning and planning of top-down
shifts of attention, and of action
\cite{Navalpakkam_Itti05vr,Itti_Arbib06arl}:

\begin{itemize}
\item {\bf Interpret task definition:} by evaluating the relevance of
  known entities (in long-term symbolic memory) to the task at hand,
  and storing the few most relevant entities into symbolic working
  memory. For example, if the task is to drive, be alert to traffic
  signs, pedestrians, and other vehicles.

\item {\bf Prime visual analysis:} by priming spatial locations that
  have been learned to usually be relevant, given a set of desired
  entities and a rapid analysis of the ``gist'' and rough layout of
  the environment \cite{Rensink00,Torralba03}, and by priming the
  visual features (e.g., color, size) of the most relevant entities
  being looked for \cite{Wolfe94}.

\item {\bf Attend and recognize:} the most salient location given the
  priming and biasing done at the previous step. Evaluate how the
  recognized entity relates to the relevant entities in working
  memory, using long-term knowledge of inter-relationships among
  entities.

\item {\bf Update:} Based on the relevance of the recognized entity,
  decide whether it should be dropped as uninteresting or retained in
  working memory (possibly creating an associated summary ``object
  file'' \cite{Kahneman_etal92} in working memory) as a potential
  object and location of interest for action planning.

\item {\bf Iterate:} the process until sufficient information has been
  gathered to allow a confident decision for action.

\item {\bf Act:} based on the current understanding of the visual
  environment and the high-level goals.

\end{itemize}

An example top-down model that includes the above elements, although
not in a very detailed implementation, was proposed by Navalpakkam \&
Itti \shortcite{Navalpakkam_Itti05vr}. Given a task definition as
keywords, the model first determines and stores the task-relevant
entities in symbolic working memory, using prior knowledge from
symbolic long-term memory. The model then biases its saliency-based
attention system to emphasize the learned visual features of the most
relevant entity. Next, it attends to the most salient location in the
scene, and attempts to recognize the attended object through
hierarchical matching against stored representations in visual
long-term memory. The task-relevance of the recognized entity is
computed and used to update the symbolic working memory. In addition,
a visual working memory in the form of a topographic task-relevance
map (TRM) is updated with the location and relevance of the recognized
entity.  The implemented prototype of this model has emphasized four
aspects of biological vision: determining task-relevance of an entity,
biasing attention for the low-level visual features of desired
targets, recognizing these targets using the same low-level features,
and incrementally building a visual map of task-relevance. The model
was tested on three types of tasks: single-target detection in 343
natural and synthetic images, where biasing for the target accelerated
its detection over two-fold on average; sequential multiple-target
detection in 28 natural images, where biasing, recognition and working
memory contributed to rapidly finding all targets; and learning a map
of likely locations of cars from a video clip filmed while driving on
a highway \cite{Navalpakkam_Itti05vr}.

\begin{figure}[p]
\includegraphics[width=\linewidth]{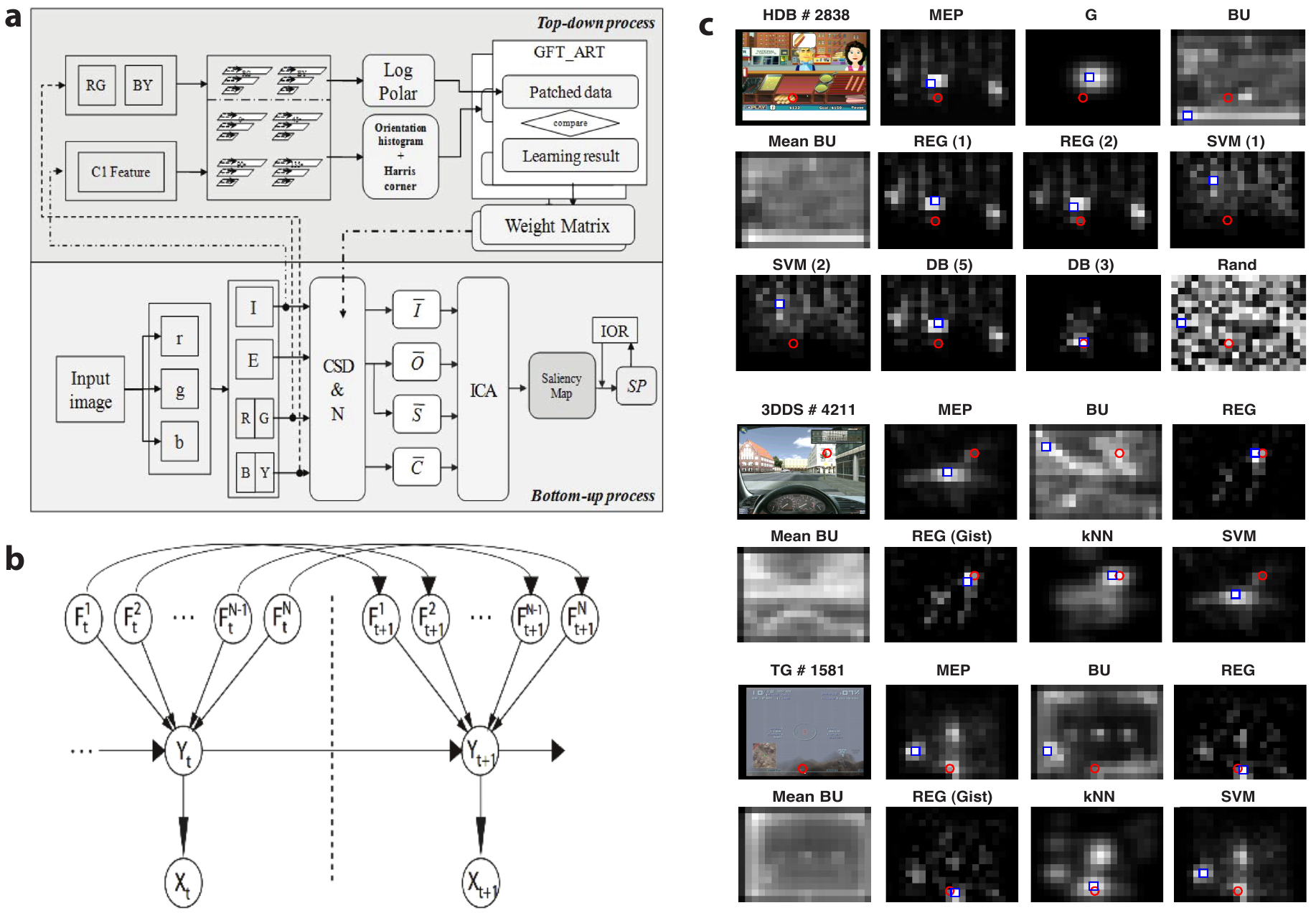}
\caption{Examples of recent top-down models. {\bf (a)} Model of Ban
  {\em et al.} (2010) which integrates bottom-up and top-down
  components. r, g, b: red, green and blue color channels. I:
  intensity feature. E: edges. R, G: red-green color. B, Y:
  blue-yellow color. CSD\&N: center surround differences and
  normalization. ICA: independent component analysis. GFT\_ART:
  growing fuzzy topology adaptive resonnance theory. SP: saliency
  point.  {\bf (b)} Graphical representation of the Dynamic Bayesian
  Network (DBN) approach of Borji {\em et al.} (2012) unrolled over
  two time-slices. $X_{t}$ is the current saccade position, $Y_{t}$ is
  the currently attended object, and $F^{i}_{t}$ is the function that
  describe object $i$ at the current scene. All variables are
  discrete. It also shows a time series plot of probability of objects
  being attended and a sample frame with tagged objects and eye
  fixation overlaid.  {\bf (c)} Sample predicted saccade maps of the
  DBN model (shown in b) on three video games and tasks: running a
  hot-dog stand (HDB; top three rows), driving (3DDS; middle two
  rows), and flight combat (TG, bottom two rows). Each red circle
  indicates the observer’s eye position superimposed with each map’s
  peak location (blue squares). Smaller distance indicates better
  prediction.  Models compared are as follows. MEP: mean eye position
  over all frames during the game play (control model). G: trivial
  Gaussian map at the image center.  BU: bottom-up saliency map of the
  Itti model. Mean BU: average saliency maps over all video
  frames. REG(1): regression model which maps the previous attended
  object to the current attended object and fixation location. REG(2):
  similar to REG(1) but the input vector consists of the available
  objects at the scene augmented with the previously attended
  object. SVM(1): and SVM(2) correspond to REG(1) and REG(2) but using
  an SVM classifier.  Similarly, DB(5) and DB(3) correspond to REG(1)
  and REG(2) meaning that in DB(5) the network considers just one
  previously attended object, while in DB(3) each network slice
  consists of the previously attended object as well information of
  the previous objects in the scene. REG(Gist): regression based only
  on the gist of the scene. kNN: k-nearest-neighbors classifier. Rand:
  white noise random map (control). Overall, DB(3) performed best at
  predicting where the player would look next (Borji {\em et al.},
  2012).}
\label{FIGtdcomplex}
\end{figure}

While the previous example model uses explicit cognitive reasoning
about world entities and their relationships, a complementary trend in
top-down modeling uses fuzzy (as in fuzzy set theory \cite{Zadeh65})
or probabilistic reasoning to explore how several sources of bottom-up
and top-down information may combine. For example, Ban {\em et al.}
\shortcite{Ban_etal10} proposed a model where the bottom-up and
top-down components interact through a fuzzy learning system
(Figure~\ref{FIGtdcomplex}.a). During training, a bottom-up saliency
map selects locations, and their features are incrementally clustered
and learned in a growing fuzzy topology adaptive reasonance theory
model (GFTART), which is a neural network model that automatically
learns to categorize many received pattern exemplars into a small (but
possibly growing) set of categories. During testing, top-down interest
in a given object activates its features stored in the GFTART model,
and biases the bottom-up saliency model to become more sensitive to
these features, thereby increasing the probability that the object of
interest will stand out. In a related approach, Akamine {\em et al.}
\shortcite{Akamine_etal12} (also see \cite{Kimura_etal08}) developed a
dynamic Bayesian network that combines the following factors: First,
input video frames give rise to deterministic saliency maps. These are
converted into stochastic saliency maps via a random process that
affects the shape of salient blobs over time (e.g., dynamic Markov
random field \cite{Kimura_etal08}). An eye focusing map is then
created which highlights maxima in the stochastic saliency map,
additionally integrating top-down influences from an eye movement
pattern (a stochastic selection between passive and active state with
a learned transition probability matrix). The authors use a particle
filter with Markov chain Monte-Carlo (MCMC) sampling to estimate the
parameters; this technique often used in machine learning allows for
fast and efficient estimation of unknown probability density
functions. Several additional recent related models using graphical
models have been proposed (e.g., \cite{Chikkerur_etal10}).

In a recent example, using probabilistic reasoning and inference
tools, Borji~{\em et al.}~\cite{borji2012object,borji2014look} introduced a
framework for top-down overt visual attention based on reasoning, in a
task-dependent manner, about objects present in the scene and about
previous eye movements.  They designed a Dynamic Bayesian Network
(DBN) that infers future probability distributions over attended
objects and spatial locations from past observed data.  Briefly, the
Bayesian network is defined over object variables that matter for the
task. For example, in a video game where one runs a hot-dog stand and
has to serve multiple customers while managing the grill, those
include raw sausages, cooked sausages, buns, ketchup, etc. Then,
existing objects in the scene, as well as the previous attended
object, provide evidence toward the next attended object
(Figure~\ref{FIGtdcomplex}.b). The model also allows to read out which
spatial location will be attended, thus allowing one to verify its
accuracy against the next actual fixation of the human player. The
parameters of the network are learned from training data in the same
form as the test data (human players playing the game). This
object-based model was significantly more predictive of eye fixations,
compared to simpler classifier-based models, several state-of-the-art
bottom-up saliency models, and control algorithms such as mean eye
position (Figure~\ref{FIGtdcomplex}.c). This points toward the
efficacy of this class of models to capture spatio-temporal
visually-guided behavior in the presence of a task.

While fully-computational top-down models are more complex than their
bottom-up counterparts, many recent examples thus exist that provide
an inspiration for future efforts in developing models that more
accurately emulate the human cognitive processes that control top-down
attention.

\section{Outlook} \label{SECdisc}

Our review shows that tremendous progress has been made in modeling
both bottom-up and top-down aspects of attention computationally. Tens
of new models have been developed, each bringing new insights into the
question of what makes some stimuli more important to visual observers
than other stimuli.

While many models have approached the problem of modeling top-down
attention, a fully implemented cognitive system that reasons about
objects, their relationships, and their locations to guide the next
shift of attention remains an elusive goal to date.

Several barriers exist in building even more sophisticated visual
attention models, which, we argue, depend on progress in complementary
aspects of machine vision, knowledge representation, and artificial
intelligence, to support some of the components required to implement
attention-driven scene understanding systems. Of prime importance is
object recognition, which remains a hard problem in machine vision,
but is necessary to enable reasoning about which object to look for
next (using top-down strategies) given the set of objects that have
been attended to and recognized so far. Also important is
understanding the spatial and temporal structure of a scene, so that
reasoning about objects and locations in space and time can be
exploited to guide attention (e.g., understanding pointing gestures,
or trajectories of objects in three dimensions). Additionally,
building knowledge bases that can capture what an observer may know
about different world entities and that allows reasoning over this
knowledge is required to build more able top-down attention
models. For example, when making tea \cite{Land_Hayhoe01}, knowledge
about different objects relevant to the task, where they usually are
stored in a kitchen, and how to manipulate them is needed to decide
where to look and what to do next.

\subsubsection*{Acknowledgements} Supported by the National Science
Foundation (grant numbers CMMI-1235539), the Army Research Office
(W911NF-11-1-0046 and W911NF-12-1-0433), the U.S. Army
(W81XWH-10-2-0076), and Google. The authors affirm that the views
expressed herein are solely their own, and do not represent the views
of the United States government or any agency thereof.

\nocite{Treisman_Gelade80,Treisman_Souther85,Milanese_etal94}

\nocite{Borji_Itti12pami,Borji_etal12tip}

\nocite{salah2002selective,gao2004discriminant,ouerhani2003real,boccignone2004modelling,ma2005generic,zhai2006visual,harel2007graph,le2006coherent,walther2006modeling,shic2007behavioral,mancas2007computational,guo2008spatio,pang2008stochastic,kootstra2008paying,ban2008dynamic,rajashekar2008gaffe,aziz2008fast,kienzle2007nonparametric,rosin2009simple,li2010visual,garcia2009decorrelation,zhang2009sunday,achanta2009frequency,gao2009discriminant,mahadevan2010spatiotemporal,li2010optimol,murray2011saliency,mccallum1996reinforcement,rao1996modeling,ramstrom2002visual,renninger2005information,jodogne2007closed,butko2009optimal,verma2009generating}

\bibliographystyle{authordate1}
\bibliography{bibliography/ilab,reviewrefs}

\end{document}